\documentclass[runningheads]{llncs}
\usepackage[T1]{fontenc}
\usepackage{graphicx}

\usepackage{amsmath}
\usepackage{microtype}
\usepackage{booktabs}
\usepackage{makecell}
\usepackage{array}
\usepackage{xurl}
\usepackage{subcaption}
\usepackage{hyperref}
\begin{document}
\sloppy

\title{Constructing Evaluation Datasets for Procedural Reasoning: Balancing Naturalness, Grounding, and Multi-Hop Coverage}

\titlerunning{Evaluation Datasets for Procedural Reasoning}

%
%
\author{Sarah Elshabrawy\and
Rahul K. Dass\and
Ashok K. Goel}
\authorrunning{S. Elshabrawy et al.}
%
\institute{Georgia Institute of Technology, Atlanta GA, USA\\
\email{\{selshabrawy3,rdass7,ag25\}@gatech.edu}}
\maketitle 

\begin{abstract}
Evaluating procedural reasoning in AI-supported learning systems requires
question-answer datasets that are both learner-like and grounded in the
instructional knowledge the system is expected to use. We study how TMK-based
question generation strategies affect dataset quality for procedural and
multi-hop reasoning.

We compare three strategies: strict generation from Task--Method--Knowledge
(TMK) models, transcript-first generation with post-hoc TMK filtering, and
TMK-aware generation that combines transcripts with structured guidance. To
evaluate generated items, we introduce a grounding validation framework based on
closed-set evidence units extracted from TMK models. The framework measures
whether answers are supported by the underlying representation, whether
questions are self-contained, and whether they target multi-hop procedural
reasoning.

Across 23 instructional topics and 690 generated question-answer pairs, strict
TMK generation achieves the strongest overall quality, with 96.5\% grounded
questions and 92.6\% usable questions. Transcript-first generation produces more
learner-like questions but more context-dependent or weakly grounded items,
while TMK-aware generation yields high raw multi-hop coverage but lower
grounding. These results show that procedural richness and natural phrasing do
not guarantee representational grounding, motivating explicit
representation-aware validation for evaluation datasets in AI-supported learning.
\keywords{evaluation datasets \and procedural reasoning \and multi-hop reasoning \and question generation \and grounding \and structured knowledge representations}
\end{abstract}
\noindent\textbf{Code and artifacts.}
Prompt templates, validation scripts, aggregation code, aggregate results,
and generated QA artifacts are available in the project repository:
\href{https://github.com/DILab-Ivy/tmk-procedural-qa-eval}{DILab-Ivy/tmk-procedural-qa-eval}.
Restricted course materials are excluded.

\section{Introduction}

AI-supported learning systems increasingly use generative and hybrid
knowledge-based AI approaches to answer student questions, provide explanations,
and support reflection 
\cite{dass2025ivy,lum2025designing}. However, evaluating whether such systems
can reason procedurally remains difficult. Many question-answer datasets test
whether a model can produce a correct or plausible answer, but do not test
whether that answer is faithful to the structured instructional knowledge the
system is expected to use.

Procedural reasoning is especially challenging because it requires connecting
multiple steps, constraints, goals, and domain concepts. In learning settings, a
good answer should not only state what is correct, but also explain how a
process works, why a step matters, and how different parts of the instructional
representation support the answer.

This paper investigates how Task--Method--Knowledge (TMK) models
\cite{chandrasekaran1992task,murdock2008meta} can be used to construct
evaluation question-answer pairs for procedural and multi-hop reasoning. We use
TMK models as structured sources of procedural knowledge and compare three
strategies for deriving questions from TMK and lesson transcripts.
We ask: \textbf{RQ1} How do generation strategies affect grounding quality?
\textbf{RQ2} How often are generated questions self-contained rather than
dependent on hidden transcript context? \textbf{RQ3} Which strategy best
balances grounding, self-containedness, and multi-hop procedural coverage?

This paper makes three contributions. First, we compare strict TMK generation,
transcript-first generation with TMK filtering, and TMK-aware generation for
constructing procedural reasoning evaluation data. Second, we introduce a
closed-evidence grounding validation framework for generated question-answer
pairs. Third, we evaluate 690 generated items across 23 instructional topics and
identify recurring failure modes in grounding, self-containedness, and
multi-hop labeling.

\section{Background and Related Work}

Our work connects three areas: structured procedural knowledge for learning,
multi-hop and procedural reasoning benchmarks, and grounding-based validation
of generated question-answer pairs.

\subsection{Structured Procedural Knowledge for Learning}

Knowledge-based learning systems have long used explicit representations of
domain knowledge, skills, and problem-solving procedures. Work on generic tasks
and task-structure analysis argued that expertise should be modeled not only as
facts, but also as goals, procedures, and the knowledge needed to support them
\cite{chandrasekaran1986generic,chandrasekaran1992task}. This view is closely
related to intelligent tutoring systems, where explicit models of knowledge
components and problem-solving steps support feedback and learning
\cite{koedinger2006cognitive,vanlehn2006behavior}.

TMK models represent procedural skills through tasks, methods, and knowledge:
tasks define goals and success or failure conditions, methods describe
procedures, and knowledge grounds those procedures in concepts, relations, and assertions.

Prior work has used TMK and related representations to model agent behavior and ground AI coaching systems for procedural explanation
\cite{murdock2008meta,goel2017gaia,dass2025ivy,lum2025designing}.

Recent text-to-model work also studies how LLMs can draft TMK models from
instructional materials for expert refinement \cite{dass2026developing}.

In contrast, we use TMK models not as a runtime tutoring substrate or authoring
target, but as structured ground truth for constructing and validating
evaluation questions.

\subsection{Evaluating Multi-hop and Procedural Reasoning}

Multi-hop question-answering benchmarks such as HotpotQA, MuSiQue, and QASC
show that dataset construction choices affect whether a benchmark actually tests
multi-step reasoning rather than shortcut retrieval
\cite{yang2018hotpotqa,trivedi2022musique,khot2020qasc}. More recent
procedural benchmarks, including ProcBench and PKR-QA, move beyond isolated
facts toward ordered steps, dependencies, and procedural knowledge graphs
\cite{fujisawa2024procbench,nguyen2026pkrqa}.

Our work is complementary: rather than proposing a new model-performance
benchmark alone, we study the construction process for procedural evaluation
data. In learning systems, a question can be natural and pedagogically plausible
but still unsuitable if its answer depends on information outside the target
instructional representation.

\subsection{Grounding and Evidence-Based Validation}

LLMs can generate fluent outputs that are unsupported or only partially
supported by their sources \cite{gao2025llm,zhang2025sirens}. Prior work on
attribution and factual consistency evaluates whether generated answers are
supported by explicit evidence, including AIS, TRUE, FActScore, and attributed
question answering
\cite{rashkin2023ais,honovich2022true,min2023factscore,bohnet2022attributedqa,hu2025caqa}.

We adapt this source-based view to procedural dataset construction. Instead of
validating answers against open-ended documents or retrieved passages, we
validate question-answer pairs against closed evidence units extracted from a
structured procedural model. This lets us distinguish natural transcript-like
questions from questions whose answers are actually supported by the intended
representation.

\section{Method}
\label{sec:method}

\begin{figure}[t]
    \centering
    \includegraphics[width=\linewidth]{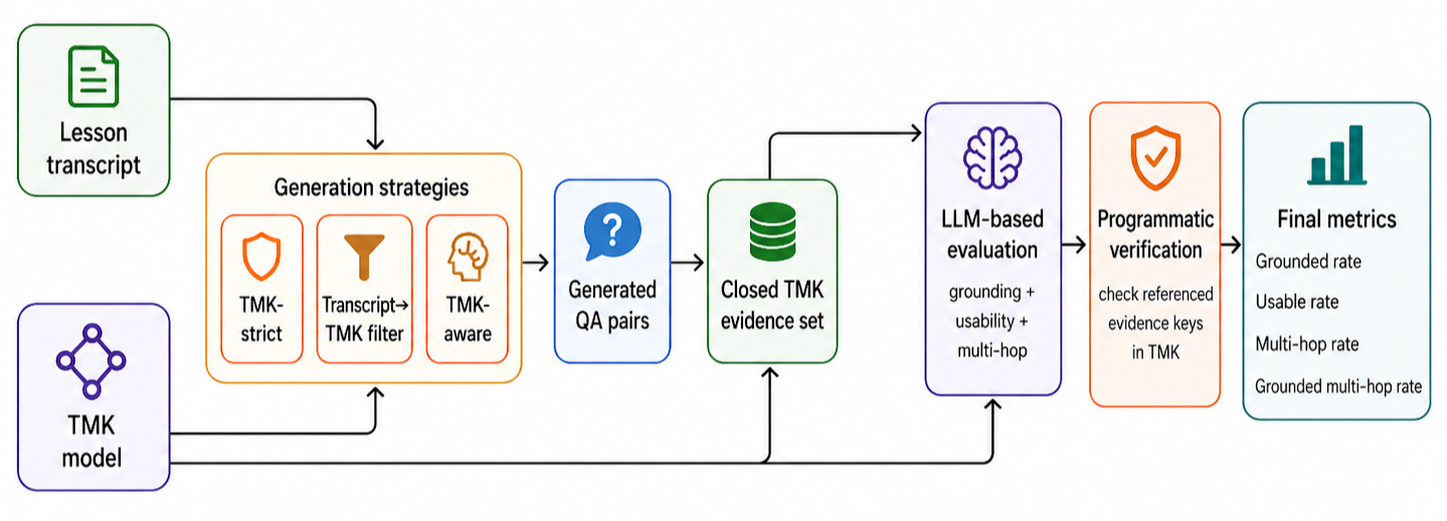}
   \caption{Question-answer generation and validation pipeline. Transcripts and TMK
models are used to generate candidate items under three strategies. Items are
validated against a closed TMK evidence set, followed by a programmatic evidence
membership check and aggregate metric computation.}
    \label{fig:pipeline}
\end{figure}

Figure~\ref{fig:pipeline} summarizes the dataset construction and validation
pipeline. For each instructional topic, we use a lesson transcript and a
corresponding TMK model. The transcript provides learner-facing phrasing and
examples, while the TMK model provides the structured procedural representation
used for generation guidance and grounding validation.

\subsection{Generation Strategies}

We compare three strategies. \textbf{TMK-aware generation} uses both the
transcript and TMK model during generation: the transcript supports natural
student-like wording, while TMK provides procedural structure. \textbf{Strict
TMK generation} treats TMK as the primary source of truth and uses the
transcript only for wording; because TMK is encoded as structured JSON with
formal task, method, state, transition, and condition fields, this strategy may
produce more schematic questions. \textbf{Transcript-first generation with TMK
filtering} first generates natural questions from the transcript, then keeps or
rewrites only items whose answers are supported by TMK.

A pilot analysis showed that some generated items were grounded but weak as
benchmark questions because they relied on hidden classroom context or
over-labeled questions as multi-hop. We therefore refined the prompts to require
self-contained wording and conservative reasoning-type labels. The final results
use these refined prompts.

\subsection{Grounding Validation}

To validate grounding, we extract a closed set of evidence units from each TMK
model. Each unit corresponds to a field in a task, method, concept, instance,
relation, assertion, or property. The validator classifies each question-answer
pair as \emph{grounded}, \emph{partially grounded}, or \emph{unsupported}. We
use \emph{unsupported} to refer to items whose answers are not grounded in the
closed TMK evidence set. For each item, the validator also selects supporting
evidence units from this closed set. We then verify programmatically that
selected evidence identifiers belong to the closed evidence set. This verifies
evidence membership, but not evidence sufficiency; we return to this limitation
in Section~\ref{sec:limitations}.

\subsection{Experimental Setup and Metrics}

We evaluate 23 topics from Georgia Tech's Knowledge-Based AI course, including
classification, planning, semantic networks, frames, constraint propagation,
case-based reasoning, diagnosis, scripts, production systems, version spaces,
commonsense reasoning, and explanation-based learning. For each topic, we use a
lesson transcript drawn from the course materials and a corresponding TMK model.

For each topic and each of the three generation strategies, we generate 10
question-answer pairs, yielding 690 total items.

Each item is evaluated for grounding, self-containedness, and reasoning type. A
question is self-contained if it can be understood without the transcript,
slides, previous questions, or classroom discussion. A question is multi-hop only
if answering it requires connecting at least two distinct pieces of evidence,
such as procedural steps, constraints, concepts, or representations. We define an
item as usable when it is both grounded and self-contained:
\[
\mathrm{Usable}(q,a)=\mathrm{Grounded}(q,a)\wedge\mathrm{SelfContained}(q).
\]
We report grounded, partially grounded, unsupported, self-contained, usable,
multi-hop, grounded multi-hop, and usable multi-hop rates.

\section{Results and Discussion}
\label{sec:results}

We report results across 690 generated question-answer pairs, with 230 items per
generation strategy. All percentages in this section are computed over the 230
items generated by each strategy unless otherwise stated.

\subsection{Overall Dataset Quality}

\begin{figure}[t]
\centering

\begin{subfigure}[t]{0.48\linewidth}
    \centering
    \includegraphics[width=\linewidth]{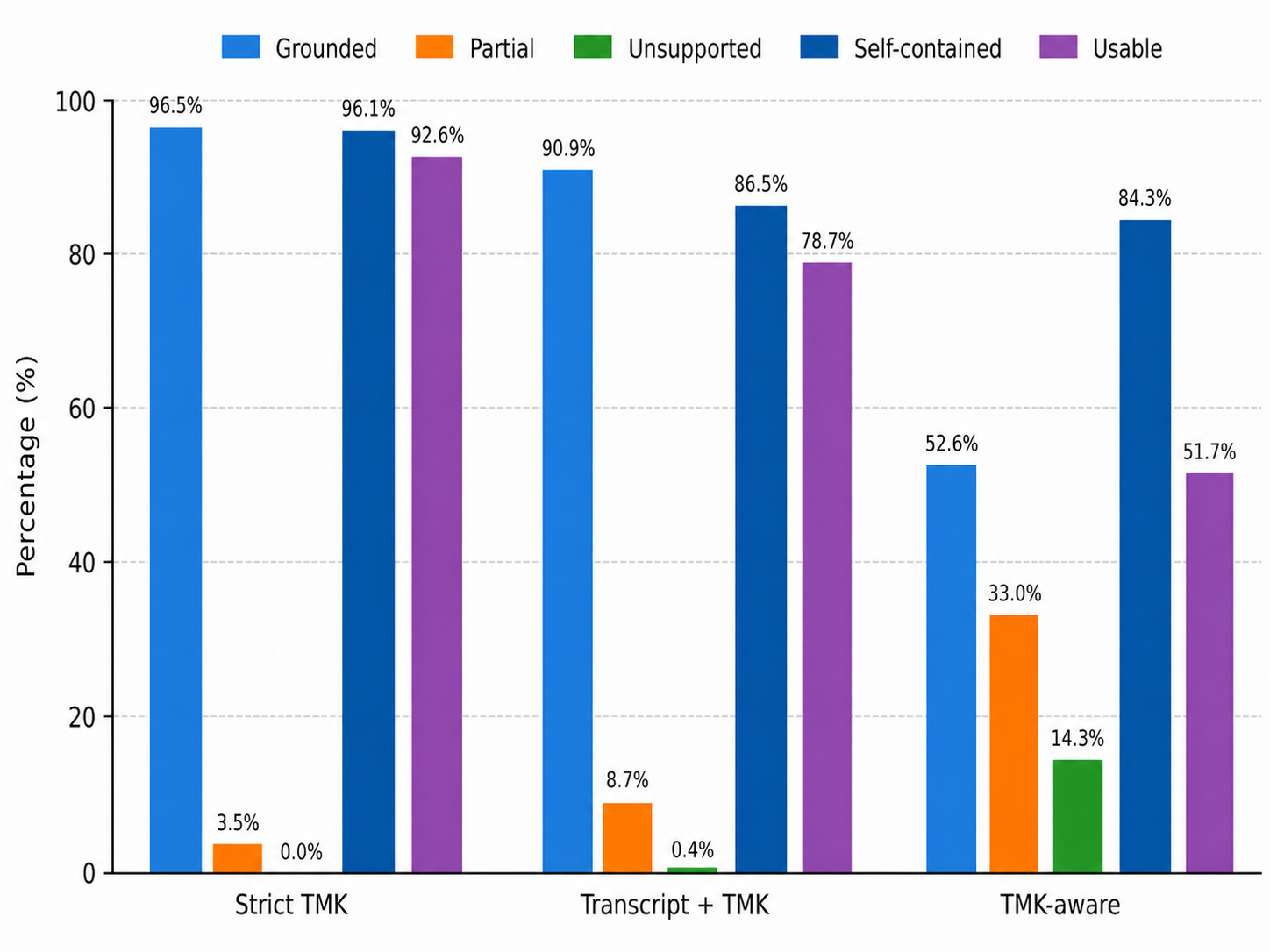}
    \caption{Overall quality.}
    \label{fig:overall-quality}
\end{subfigure}
\hfill
\begin{subfigure}[t]{0.48\linewidth}
    \centering
    \includegraphics[width=\linewidth]{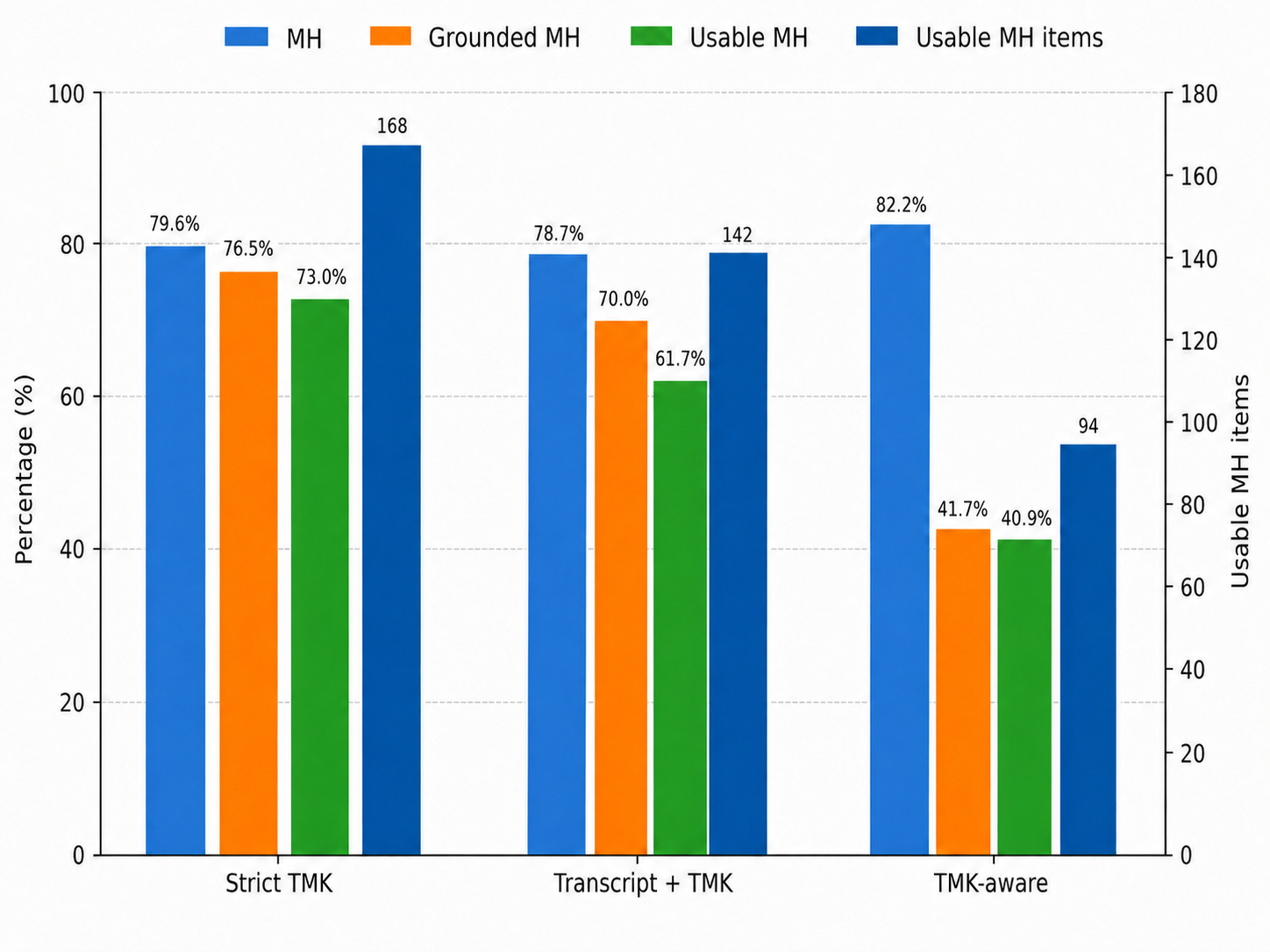}
    \caption{Multi-hop coverage.}
    \label{fig:multihop-quality}
\end{subfigure}

\caption{Quality and multi-hop coverage by generation strategy.}
\label{fig:results-summary}
\end{figure}

Figure~\ref{fig:overall-quality} summarizes grounding, self-containedness, and
usability. Strict TMK generation performs best overall, producing the highest
grounding rate, self-contained rate, and usable rate. It generated no unsupported items and only 8 partially grounded items
(3.5\% of 230).

Transcript-first generation performs second-best: post-hoc TMK filtering
recovers much of the grounding quality, but context-dependent phrasing and
unsupported transcript details remain. TMK-aware generation performs weakest on
grounding, suggesting that combining transcripts and TMK during generation can
lead the model to blend supported structure with unsupported transcript-only
details.

\subsection{Multi-Hop Procedural Coverage}

Figure~\ref{fig:multihop-quality} shows that raw multi-hop coverage can be
misleading. TMK-aware generation has the highest raw multi-hop rate, but its
usable multi-hop rate drops sharply because many items are not grounded or
self-contained. Strict TMK generation has slightly lower raw multi-hop coverage
but the highest usable multi-hop rate, producing 168 usable multi-hop items.
This makes it the strongest strategy for evaluation data that is both
procedurally rich and representationally grounded.

\subsection{Naturalness, Self-Containedness, and Grounding}

Self-containedness is a major source of quality loss. A question may be grounded
but still weak as a benchmark item if it depends on hidden lesson context, such
as references to ``the example'' or ``the process we discussed.'' The
transcript-first strategy produced the most non-self-contained questions,
showing that learner-like phrasing can preserve classroom context that is
unsuitable for standalone evaluation.

At the same time, strict TMK generation should not be interpreted as the most
natural strategy. Because it is anchored in formal JSON structures of tasks,
methods, states, transitions, and conditions, it can produce more schematic
questions. 
For example, a strict TMK item for Classification asked:
``What sequence of successful checks must occur for the overall animal-to-bird
classification process to finish successfully? Describe each step's required
condition in order.'' Although grounded, the wording mirrors the procedural
control structure of the TMK model rather than the way a learner would typically
ask about classification. A more learner-like transcript-first item asked:
``Can you walk me through the steps the agent actually takes to go from the
observed animals to a finished list of birds, and what has to be true to move
past each step?''

The central tradeoff is therefore not between good and bad
generation, but between natural student-like language and faithful alignment
with the structured representation.

\subsection{Qualitative Failure Modes}

Table~\ref{tab:qualitative-failures} shows representative validation failures
from the TMK-aware strategy. These examples illustrate why raw procedural
richness was not sufficient for benchmark quality. In several cases, generated
items appeared to require multi-step reasoning, but the answer relied on
information that was not fully available in the TMK evidence set or depended on
context from a previous question.

\begin{table}[t]
\centering
\caption{Representative failure modes from TMK-aware generation.}
\label{tab:qualitative-failures}
\small
\setlength{\tabcolsep}{4pt}
\renewcommand{\arraystretch}{1.15}
\begin{tabular}{p{2.5cm}p{4.2cm}p{5.3cm}}
\toprule
\textbf{Failure mode} &
\textbf{Example pattern} &
\textbf{Why validation rejects it} \\
\midrule

Unsupported causal links &
The answer infers causal rules from examples even though the TMK model only
lists associated features or properties. &
The selected evidence shows feature co-occurrence, but not an explicit
implication or causal relation. The item therefore treats association as causal
support. \\

Hidden prior context &
The question assumes that earlier steps, intermediate results, or a previous
scenario have already been established. &
The item may be understandable in a lesson sequence, but it is not
self-contained as a standalone benchmark question. The validator marks it for
rewrite rather than direct use. \\

Transcript-only detail &
The question or answer uses concrete examples, rules, or explanatory details
from the transcript that are not represented in the TMK evidence set. &
The answer may follow the intended procedure, but some proof steps depend on
information outside the closed TMK evidence set. This makes the item only
partially grounded. \\

\bottomrule
\end{tabular}
\end{table}

\subsection{Implications}

These results suggest that procedural evaluation datasets should report
combined metrics such as grounded multi-hop and usable multi-hop rates, rather
than raw multi-hop coverage alone. For AI-supported learning interfaces,
evaluation questions should be validated against the same instructional
representation the system is expected to use; otherwise, evaluation may reward
fluent answers to questions outside the modeled knowledge.

\section{Limitations and Future Work}
\label{sec:limitations}

This study has several limitations. First, grounding validation still relies on
LLM judgment. Although the validator is constrained by closed TMK evidence
units and we programmatically verify evidence membership, we do not
independently verify evidence sufficiency. Second, we do not include full manual
expert annotation of all generated items, which would provide a stronger gold
standard for grounding and self-containedness. Third, our current analysis
reports descriptive aggregate rates by generation strategy and does not model
the nesting of generated items within instructional topics; future analyses
should estimate strategy effects using hierarchical models that account for
topic-level variation. Fourth, all topics come from one symbolic AI course, so
future work should test other domains, instructional styles, and knowledge
representations. Finally, we evaluate dataset quality rather than downstream
learner outcomes or AI tutor performance.

Future work will extend this framework toward larger-scale procedural reasoning
benchmarks in which each item is explicitly linked to structured evidence. Such
benchmarks could evaluate whether models can explain why steps are needed, how
substeps interact, what happens when constraints are violated, and how
dependencies unfold across multiple reasoning hops. Another direction is
interactive evaluation: procedural learning often involves clarification,
follow-up questions, and guided questioning, so future datasets could include
learner-like follow-ups or Socratic tutoring exchanges.

\section{Conclusion}

We compared three TMK-based strategies for constructing evaluation datasets for
procedural and multi-hop reasoning. Using closed-set evidence units extracted
from TMK models, we evaluated whether generated question-answer pairs were
grounded, self-contained, usable, and multi-hop.

Strict TMK generation produced the most reliable evaluation data, while
transcript-first generation produced more learner-like but more
context-dependent items. TMK-aware generation produced high raw multi-hop
coverage, but often blended grounded content with unsupported transcript
details. These findings show that natural phrasing and procedural richness are
not sufficient for high-quality evaluation data. Evaluation datasets for
AI-supported learning should include explicit representation-aware validation so
that questions test the knowledge the system is expected to use.

\bibliographystyle{splncs04}
\bibliography{references}


\end{document}